\documentclass[10pt,twocolumn,letterpaper]{article}

\usepackage{iccv}
\usepackage{times}
\usepackage{epsfig}
\usepackage{graphicx}
\usepackage{amsmath}
\usepackage{amssymb}

\usepackage{mathtools}
\usepackage{amsthm}
\usepackage{microtype}
\usepackage{subfigure}

\usepackage{booktabs} 
\usepackage{multirow}
\usepackage{caption}

\usepackage{algorithm}
\usepackage{algorithmicx}
\usepackage{algpseudocode}

\newcommand{\name}[0]{PIAT\xspace}

\usepackage{xcolor}

\usepackage[pagebackref=true,breaklinks=true,letterpaper=true,colorlinks,bookmarks=false]{hyperref}

\iccvfinalcopy 

\def\iccvPaperID{9437} 

\ificcvfinal\fi

\begin{document}

\title{PIAT: Parameter Interpolation based Adversarial Training for Image Classification}


\author{
Kun He\thanks{The first three authors contributed equally. Correspondence to Kun He.}, Xin Liu\footnote[1]{}, Yichen Yang\footnote[1]{}\\ 
School of Computer Science and Technology, \\
Huazhong University of Science and Technology, Wuhan, China\\
{\tt\small \{brooklet60, liuxin\_jhl, yangyc\}@hust.edu.cn }
\and
Zhou Qin, Weigao Wen, Hui Xue \\
Alibaba Group, China \\
{\tt\small \{qinzhou.qinzhou, weigao.wwg, hui.xueh\}@alibaba-inc.com} 
\and
John E. Hopcroft \\
Department of Computer Science, Cornell University, Ithaca, NY, USA \\
{\tt\small  jeh@cs.cornell.edu }
}

\maketitle
\ificcvfinal\thispagestyle{empty}\fi

\begin{abstract}
Adversarial training has been demonstrated to be the most effective approach to defend against adversarial attacks. However, existing adversarial training methods show apparent oscillations and overfitting issue in the training process, degrading the defense efficacy.  In this work, we propose a novel framework, termed Parameter Interpolation based Adversarial Training (PIAT), that makes full use of the historical information during training. Specifically, at the end of each epoch, PIAT tunes the model parameters as the interpolation of the parameters of the previous and current epochs. Besides, we suggest to use the Normalized Mean Square Error (NMSE) to further improve the robustness by aligning the clean and adversarial examples. Compared with other regularization methods, NMSE focuses more on the relative magnitude of the logits rather than the absolute magnitude. Extensive experiments on several benchmark datasets and various networks show that our method could prominently improve the model robustness and reduce the generalization error. Moreover, our framework is general and could further boost the robust accuracy when combined with other adversarial training methods. 

\end{abstract}

\section{Introduction}
Deep neural networks (DNNs) have been widely used in various tasks of computer vision~\cite{Resnet18} and natural language processing~\cite{DBLP:conf/naacl/DevlinCLT19}. 
However, even the model performance surpasses humans in some tasks, they are known to be vulnerable to adversarial examples by injecting malicious and imperceptible perturbations to clean inputs that can cause the model to misclassify inputs with the high confidence~\cite{DBLP:journals/corr/SzegedyZSBEGF13,DBLP:journals/corr/GoodfellowSS14,PGD,AA,DBLP:conf/icml/AthalyeC018,DBLP:conf/icml/HendrycksLM19,DBLP:conf/iccv/XieWZZXY17,DBLP:conf/ccs/MengC17}. 
Since DNNs are applied in many safety systems, it is of great importance to make them more reliable and robust.

\begin{figure}[t]
    \centering
    \includegraphics[width=\columnwidth]{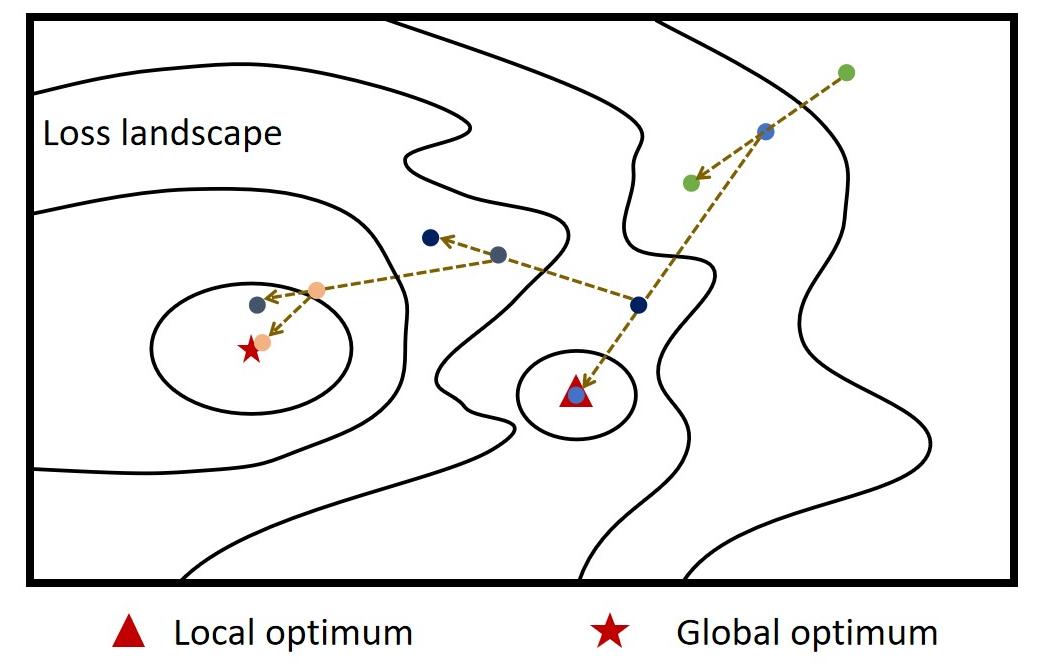}
    \caption{Illustration of the parameter update 
     for the proposed method. 
    The start and end points of each epoch are in colored dots. }
    \label{fig:MOMAT training process}
\end{figure}

To improve the model's robustness against adversarial attacks, Adversarial Training (AT) is known to be the most effective approach to defend against adversarial attacks, which generates adversarial examples during the training and incorporates them into the training.
However, the training process of AT has apparent oscillations in the early stage and overfitting issues in the later stage.
In the early stage, the model parameters are updated rapidly with a large learning rate, thus the adversarial examples generated at each epoch are pretty different, leading to the oscillation on robust accuracy.
In the later stage, many works~\cite{MLCAT,Overfitting1,EMAT} and our experiments 
have shown that the overfitting issue occurs in the later stage of AT, that is, the training accuracy continues to increase but the robust accuracy on the testing data begins to decline.

To solve the oscillation issue in the early stage of AT and the overfitting issue in the later stage, we propose a Parameter Interpolation based Adversarial Training (\name) framework. As illustrated in Figure~\ref{fig:MOMAT training process}, 
at the end of each epoch of the training, 
we tune the model parameters as the interpolation of the model parameters of the previous and current epochs.
In the early stage of AT, the adversarial examples generated at each epoch is significantly different, and the model's decision boundary changes dramatically. 
In contrast, \name tunes the model parameters with the previous epoch and makes the change of the decision boundary more moderate, 
helping to eliminate the oscillation. 
In the later stage, \name considers the previous boundary, preventing the decision boundary from becoming too complex and alleviating the overfitting issue. 
As the training continues and the model parameters become more 
valuable, \name gradually increases the weight of the previous parameters when tuning the current parameters.

On the other hand, we observe that some works, such as TRADES~\cite{Trade-off} and MART~\cite{MART}, assume that the data distribution between clean and adversarial examples are virtually indistinguishable. To improve the performance of AT, they propose regularization terms to force the model output for clean and adversarial examples to be as similar as possible. 
However, the data distribution between clean and adversarial examples is quite different, and simply forcing the output to be close is too demanding.
We suggest that AT should pay more attention to aligning the relative magnitude rather than the absolute magnitude of logits between the clean and adversarial examples. 
Hence, we propose a new metric called Normalized Mean Square Error (NMSE) to better align the clean and adversarial examples. NMSE pays more attention to the relative magnitude of the output of clean examples and adversarial examples rather than the absolute magnitude, thus learning the common information of different distributed data.

We incorporate the Normalized Mean Square Error (NMSE) as the regularization into the proposed \name framework. 
Extensive experiments on CIFAR10, CIFAR100, and SVHN datasets show that our method performs better and effectively improves the adversarial robustness of the model against white-box and black-box attacks. In addition, our \name framework is general and 
other adversarial training methods can be incorporated into our framework 
to achieve better robustness and performance.

Our main contributions are summarized as follows:
\begin{itemize}
\item To solve the oscillation issue in the early stage of AT and the overfitting issue in the later stage, we propose the \name framework that interpolates the model parameters of the previous and current epochs to consider the historical information during the training.

\item We propose to use the NMSE loss as a new regularization term to better align the logits of clean and adversarial examples. NMSE pays more attention to the relative magnitude of the output of clean examples and adversarial examples rather than the absolute magnitude.

\item  The robust accuracy of our method is stable, indicating that \name alleviates the oscillation  and overfitting issues during the AT process. Extensive experiments on three standard datasets and two networks show that \name combined with NMSE offers excellent robustness without incurring additional cost.
\end{itemize}

\section{Related Work}
\subsection{Adversarial Training} 
Adversarial training (AT)~\cite{PGD} has been demonstrated to be the most effective defensive methods against adversarial attacks, which generates a locally most adversarial perturbed point for each clean example and trains the model to classify them correctly. 

Given an image classification task, the training dataset $D=\{(\mathbf{x_i},y_i)\}_{i=1}^n$ consists of $n$ clean examples with $c$ classes, where $\mathbf{x_i} \in \mathbb{R}^d$ represents a clean example
with the ground-truth label $y_i\in \{1,2,...,c\}$. 
The adversarial training optimization problem can be formulated as the following min-max problem: 
\begin{equation}
\min_{\boldsymbol{\theta}}\sum_{i}\max_{\mathbf{x_i'}\in \mathbf{S(x_i,\epsilon)}}
\mathcal{L}(f_{\boldsymbol{\theta}}(\mathbf{x_i'},y_i)),
\end{equation}
where $f_\theta(\cdot): \mathbb{R}^d \rightarrow \mathbb{R}^c$ is the DNN classifier with parameter $\theta$. $\mathcal{L}(\cdot, \cdot)$ represents the cross entropy loss and $\mathbf{S(x_i,\epsilon)}=\{||\mathbf{x'_i}-\mathbf{x_i}||_p \leq \epsilon\}$ represents an $\epsilon$-ball of a benign data point $\mathbf{x_i}$. $\mathbf{x_i'}$ denotes the adversarial example generated from $\mathbf{x_i}$.

To solve the inner maximization problem, the adversarial example \(\mathbf{x_i'}\) is often crafted  by the Projected Gradient Decent (PGD) attack~\cite{PGD}, which can be formulated by:
\begin{equation}
    \mathbf{x}_i^{t+1}=\prod_{\mathcal{S}(\mathbf{x}_i)}(\mathbf{x}_i^{t}+\alpha \cdot sign( \nabla_{\mathbf{x}_i^t}\mathcal{L}(f_\theta(\mathbf{x}_i^{t}),y_i)).  
\end{equation}
Here $\mathbf{x}_i^t$ denotes the adversarial example at the $t^{th}$ step, $\prod(\cdot)$ is the projection operator and $\alpha$ is the step size.


TRADES~\cite{Trade-off} is another typical AT proposed to achieve a better balance between accuracy on clean examples and robustness on adversarial examples:
\begin{equation}
\begin{split}
    \min_{\boldsymbol{\theta}}\sum_{i}\{ & CE(f_{\boldsymbol{\theta}}(\mathbf{x}_i),y_i)\\&
    +\beta \cdot \max_{\mathbf{x_i'}\in \mathbf{S(x_i,\epsilon)}}KL(f_{\boldsymbol{\theta}}\mathbf({\mathbf{x}_i})||f_{\boldsymbol{\theta}}(\mathbf{x}_i')) \}, 
\end{split}
\end{equation}
where \(CE(\cdot, \cdot)\) is the cross entropy loss that 
maximize the accuracy of clean examples.  
\(KL(\cdot || \cdot)\) is the Kullback-Leibler divergence that measures the distance of output between clean and adversarial examples and $\beta$ controls the tradeoff between accuracy and robustness.

Since the inception of TRADES, subsequent efforts have been devoted to achieve better performance. 
MART~\cite{MART} explicitly differentiates the misclassified and correctly classified examples during the training. 
STAT~\cite{SqueezeAT} takes both advrsarial examples and 
collaborative examples into account for regularizing the loss landscape.
FreeAT~\cite{FreeAT} uses each epoch of PGD adversarial examples and updates the model with the gradient.
Among the adversarial data that are confidently misclassified, rather than employing the most adversarial data that maximize the loss, FAT~\cite{DBLP:conf/icml/ZhangXH0CSK20}  searches for the  least adversarial data that minimize the loss. 
LAS-AT~\cite{DBLP:conf/cvpr/JiaZW0WC22} learns to automatically produce attack strategies to improve the model robustness.
SCORE~\cite{DBLP:conf/icml/PangLYZY22} facilitates the reconciliation between robustness and accuracy, while still handling the worst-case uncertainty via robust optimization.

\subsection{Decision Boundary and Overfitting}
Dong \etal~\cite{EMAT} have found that the decision boundaries for adversarial training are more complex than those for standard training, thus making the training more difficult.
LBGAT~\cite{Learnable} 
constrains logits from the robust model that takes adversarial examples as input and makes it similar to those from the clean model fed with the corresponding natural data to better fine-tune the decision boundaries. 

Moreover, overfitting issue widely exists during the course of adversarial training~\cite{Overfitting1}.
GAIRAT~\cite{GAIRAT} adjusts the weights based on the difficulty of attacking a benign data point.
AWP~\cite{AWP} explicitly regularizes the flatness of the weight loss landscape, forming a double-perturbation mechanism in the adversarial training framework that adversarially perturbs both inputs and weights.
RLFAT~\cite{RobustLocalFeature} learns robust local features by adversarial training on the random block shuffle transformed adversarial examples, and then transfers the robust local features into the training of normal adversarial examples.
MLCAT~\cite{MLCAT} learns large-loss data as usual, and adopts additional measures to increase the loss of small-loss data to hinder data fitting when the data become easy to learn.  

Nevertheless, the classification accuracy of adversarial training on clean and adversarial examples still has room for improvement.
After each epoch of parameter updates on the model, different adversarial examples are generated. 
Compared with clean examples, adversarial examples are much more diverse, making it hard for the training to learn from the data when the network capacity is rather limited.
In this work, we observe that previous works did not fully utilize the historical information during training, thus we proposed to utilize historical information to boost the model's robustness. 

\section{Motivation}
\label{section:3}
\begin{figure*}
    \centering
    \subfigure[Test accuracy on clean examples]{
    \includegraphics[width=\columnwidth]{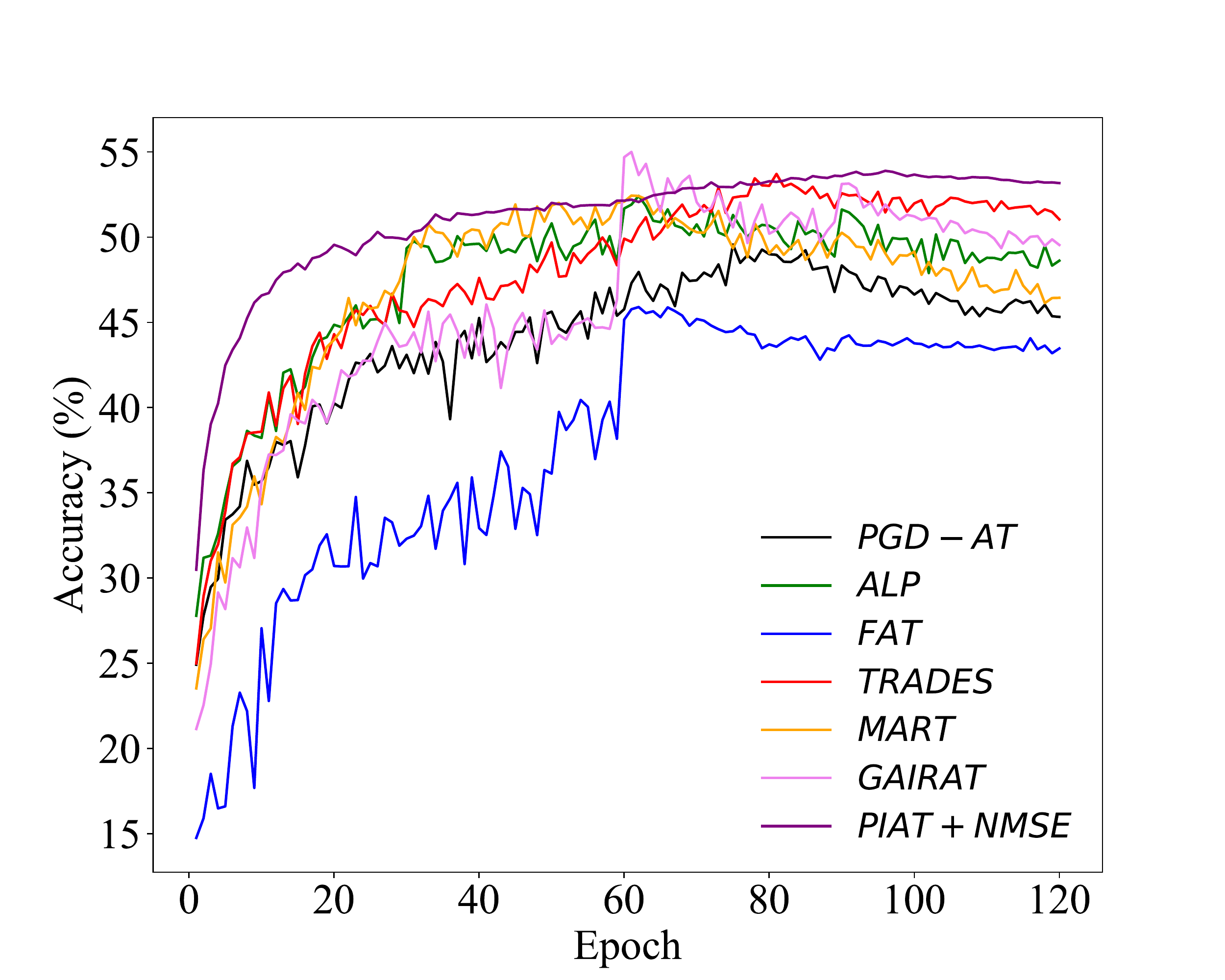}}
    \subfigure[Test accuracy on adversarial examples]{
    \includegraphics[width=\columnwidth]{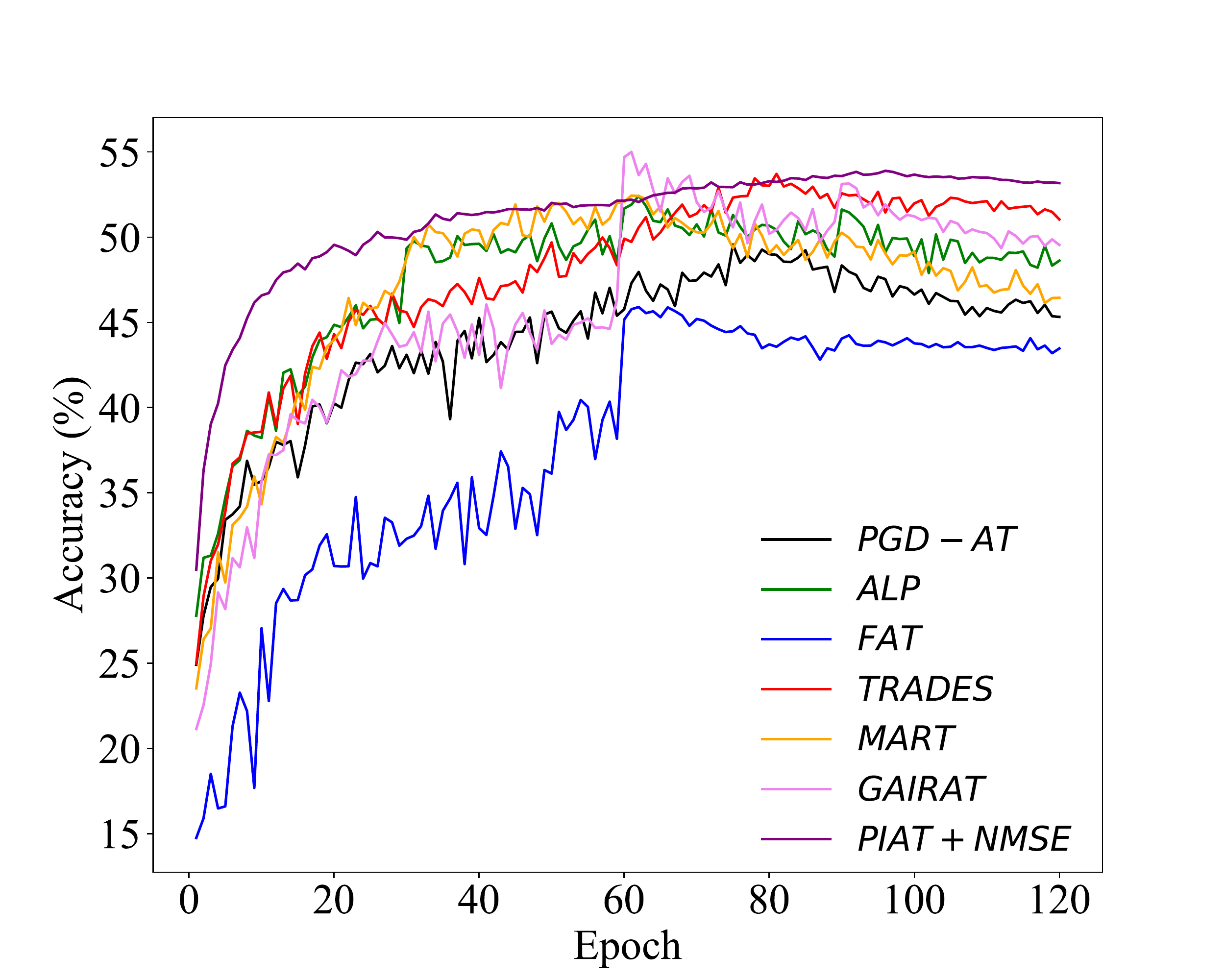}}
    \caption{
    Test accuracy on clean and adversarial examples of ResNet18 trained by various AT methods on 
    CIFAR10. }
    \label{fig:Classsical Adversarial Training}
\end{figure*}
This section further analyzes the oscillations in the early stage and the overfitting issue in the later stage of the AT process. Based on our observations, we propose to fully utilize the historical training information by parameter interpolation at the end of each epoch. We also discuss aligning clean and adversarial examples in AT and suggest focusing more on the relative magnitude of the output logits rather than the absolute magnitude.

\subsection{Analysis on the Robust Accuracy}

Figure~\ref{fig:Classsical Adversarial Training} illustrates the test accuracy on clean and adversarial examples of several popular AT methods during training, including PGD-AT~\cite{PGD}, ALP~\cite{DBLP:journals/corr/abs-1803-06373}, FAT~\cite{DBLP:conf/icml/ZhangXH0CSK20}, TRADES~\cite{Trade-off}, MART~\cite{MART} and GAIRAT~\cite{GAIRAT}. 
As illustrated in Figure~\ref{fig:Classsical Adversarial Training} (b), over all these AT methods except our method (\name + NMSE), we can observe that in the early stage, there are apparent oscillations in terms of robust accuracy, and in the later stage, the adversarial robustness all declines as the model overfits adversarial examples.

In the early stage, the model has not learned to fit the data yet and model parameters are updated with a large learning rate, leading to rapid changes on the decision boundary. Subsequently, the adversarial examples generated by the PGD attack for adversarial training vary significantly at each epoch. Thus, it is hard for the models to learn the features and have good generalization from the changing data, making the robust accuracy unstable. 
If the change on decision boundary is more moderate, then the difference of  adversarial examples generated between two adjacent epochs will be smaller.  
A straightforward solution is to reduce the learning rate. However, a low learning rate will 
decelerate the convergence of AT. Worse still, the oscillation may disappear with a low learning rate, but the overfitting issue occurs. 


In the later stage, as the learning rate of most AT methods is reduced significantly, 
the decision boundary will become overly complex when 
the model struggles to well fit the adversarial examples crafted during training. 
As a result, the learned model is not robust enough to well handle unseen adversarial examples, leading to the overfitting issue. 

We observe that existing AT methods ignore the historical information in the training process, which would be useful to stabilize the training and alleviate overfitting.
To this end, we propose to utilize the model parameters of previous epoch to fine-tune the current parameters at the end of each epoch. Such an approach allows us to leverage historical information and improve the 
the robust generalization capability. 
In the early stage, considering parameters of the previous epoch will result in a more gradual update on the model parameters, helping to ensure a smoother and more stable training. In the later stage, mixing parameters of the previous epoch could smooth the decision boundary and prevent the model from overfitting. 

\subsection{Analysis on the AT Regularization}

Combining regularization with the loss of standard AT is one of the most effective ways to alleviate the overfitting issue and improve the model robustness.
For instance, the Kullback-Leibler (KL) divergence between the classification probabilities of clean and adversarial examples is often used for regularization~\cite{Trade-off}. Minimizing the KL divergence aligns the prediction of clean and adversarial examples and encourages the output to be smooth.

However, adversarial examples contain malicious noise that 
substantially deviates from the data distribution of clean examples. We argue that it is too demanding to force the absolute magnitude of output predictions to remain the same on the two types of examples.
Moreover, the KL divergence compares the classification probabilities obtained through the softmax operation on logits. This operation tends to amplify the probability of the correct class while reducing that of other classes. AT can not fully learn the information from adversarial examples by focusing too much on the output probability of the correct class but neglecting probabilities of other classes. 

Therefore, it is more reasonable that the output logits for clean examples and adversarial examples are roughly similar but not identical. Rather than considering the absolute magnitude of the output logits, we should pay more attention to the relative magnitude between clean and adversarial examples.


\section{Methodology}
\label{sec:method}
\begin{figure*}
    \centering
    \subfigure[Test accuracy on clean examples]{
    \includegraphics[width=\columnwidth]{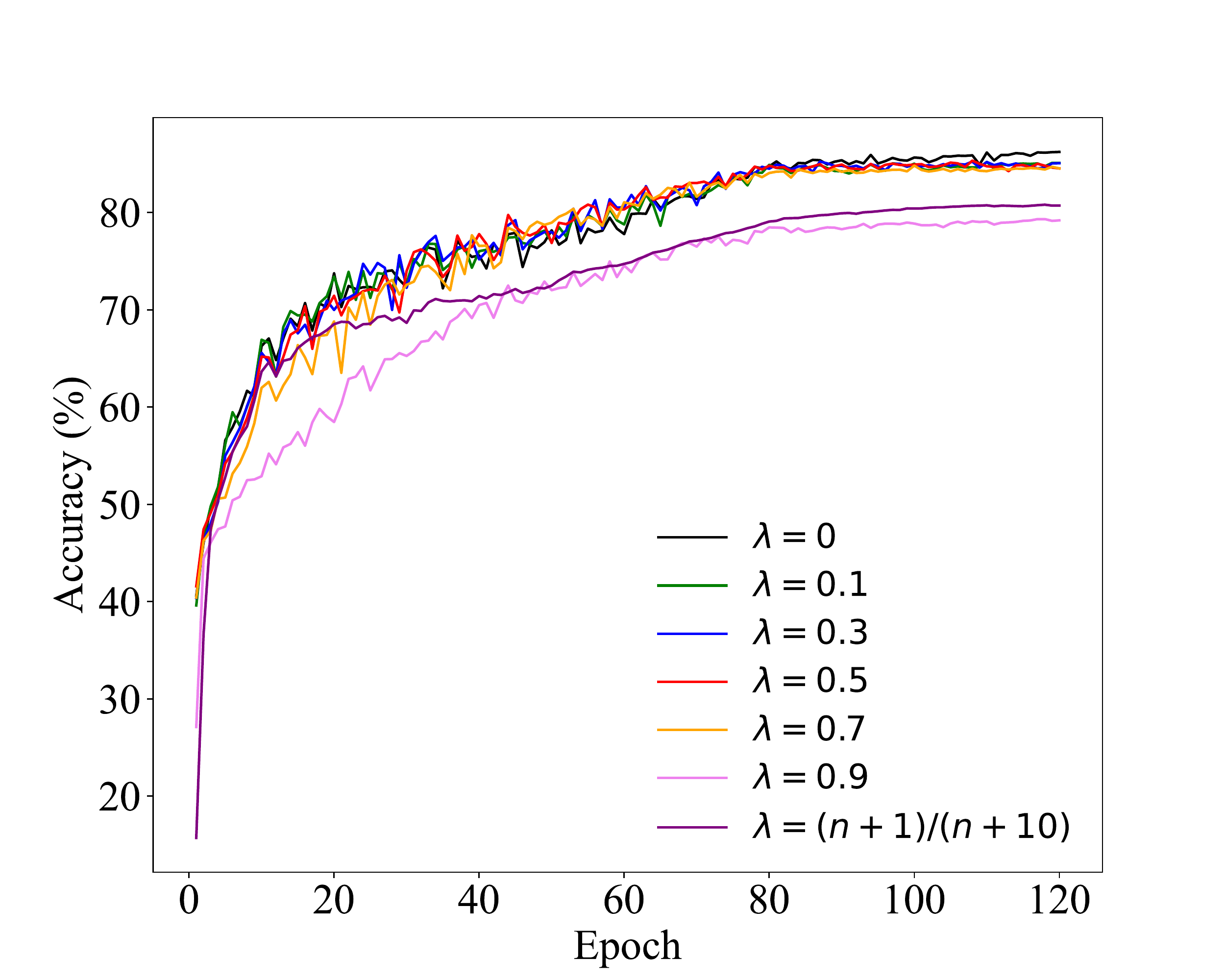}}
    \subfigure[Test accuracy on adversarial examples]{
    \includegraphics[width=\columnwidth]{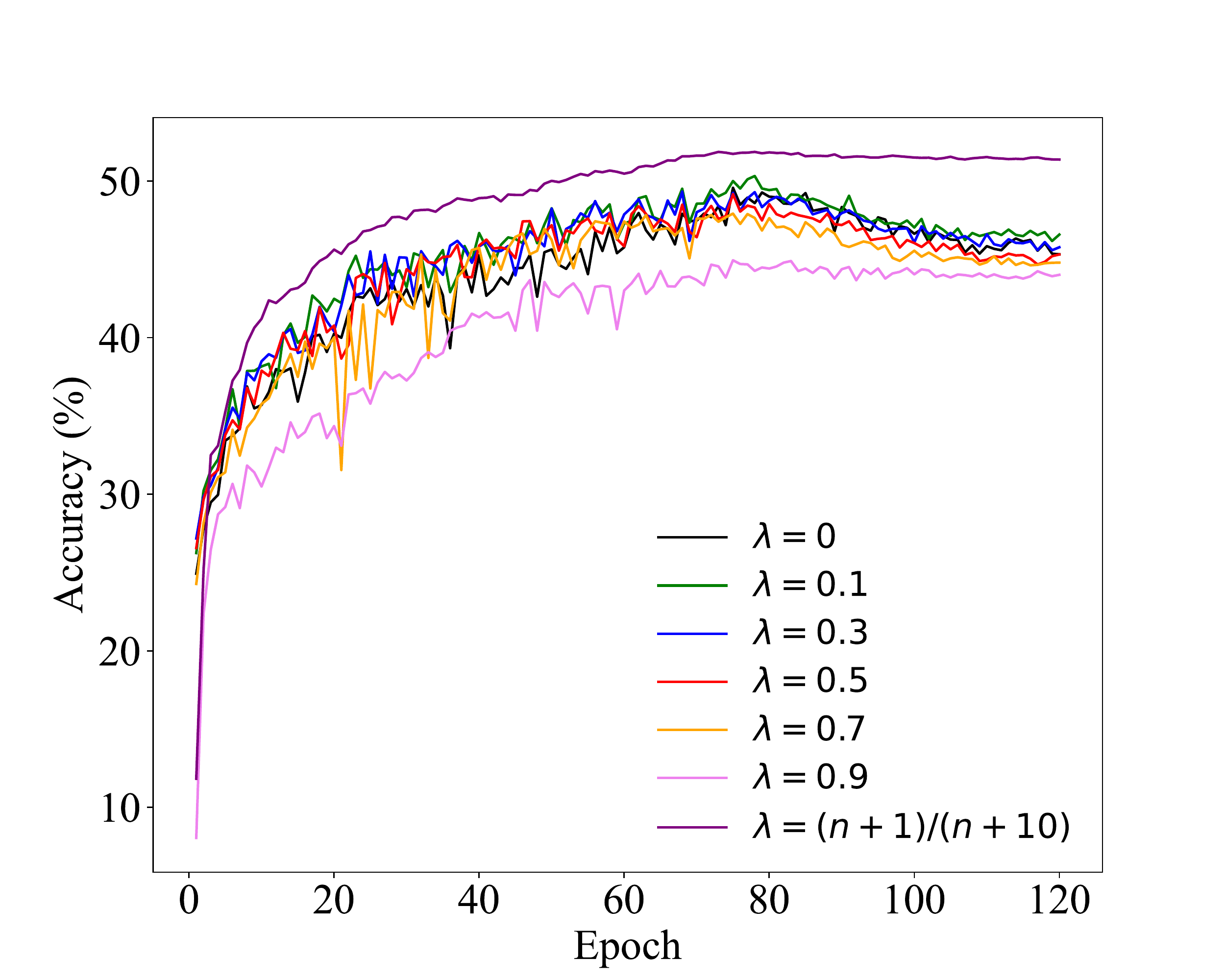}}
    \caption{
    The accuracy on clean and adversarial examples of ResNet18 models trained by \name with different \(\lambda\) on the CIFAR10 dataset. \(n\) denotes the current number of training epochs.
    }
    \label{fig:MOMAT_test_accuracy}
\end{figure*}
\begin{algorithm}[t]
    \caption{The \name Framework}
    \label{alg: MOMAT}
    \begin{algorithmic}
    \State {\bfseries Input:} Initial model parameters $\boldsymbol{\theta}_0$, perturbation size $\boldsymbol{\epsilon}$, number of adversarial attack steps \(K\), number of epochs $N, N_0$, weight function $g(\cdot)$
    \State {\bfseries Output:} $\boldsymbol{\theta}_N'$
    \State Initialize $\boldsymbol{\theta}\leftarrow \boldsymbol{\theta}_0$
    \For{$i=1$ {\bfseries to} $N_0$}
    \For{$minibatch$ $\mathbf{x}\subset\mathbf{X}$}
    \State $loss=\mathcal{L}_{CE}(\mathbf{x},y)$
    \State update $\boldsymbol{\theta}_i$
    \EndFor
    \EndFor
    \State \(\boldsymbol{\theta}_0' \leftarrow \boldsymbol{\theta}_{N_0}\)
    \For{$i=1$ {\bfseries to} $N$}
    \State $\boldsymbol{\theta}_i\leftarrow\boldsymbol{\theta}'_{i-1}$
    \For{$minibatch$ $\mathbf{x}\subset\mathbf{X}$}
    \State $\mathbf{x_{adv}}$ $\leftarrow$ $\mathbf{x}$
    \For{$k=1$ {\bfseries to} $K$}
    \State $\mathbf{x_{adv}}\leftarrow\mathbf{x_{adv}}+\boldsymbol{\epsilon}\cdot sign(\nabla_\mathbf{x}\mathcal{L}_{CE}(\mathbf{x_{adv}},y))$
    \State $\mathbf{x_{adv}}\leftarrow clip(\mathbf{x_{adv}},\mathbf{x}-\boldsymbol{\epsilon},\mathbf{x}+\boldsymbol{\epsilon})$
    \EndFor
    \State $loss=\mathcal{L}(\mathbf{x_{adv}},y)$
    \State update $\boldsymbol{\theta}_i$
    \EndFor
    \State $\lambda \leftarrow g(i)$ 
    \State $\boldsymbol{\theta}_i'\leftarrow\lambda\cdot\boldsymbol{\theta}_{i-1}'+(1-\lambda)\cdot\boldsymbol{\theta}_i$
    \EndFor
    \State \textbf{return} $\boldsymbol{\theta}_N'$
    \end{algorithmic}
\end{algorithm}
In this section, we introduce the realization of the \name framework. We also describe how to 
combine with our proposed Normalized Mean Square Error (NMSE).

\subsection{The \name Framework}
To fully utilize the historical information during training, at the end of each epoch, \name tunes the model parameters as the interpolation of parameters of the previous and current epochs, which can be 
formalized as:
\begin{equation}
    \boldsymbol{\theta_t'}=\lambda\cdot\boldsymbol{\theta_{t-1}'}+(1-\lambda)\cdot\boldsymbol{\theta_t}, \quad 0 \leq \lambda \leq 1 , 
    \label{eq:MOMAT}
\end{equation}
where \(\boldsymbol{\theta_{t-1}'}\) is the model parameters of the previous epoch after interpolation, and \(\boldsymbol{\theta_t}\) is current parameters before interpolation at the end of the training epoch.  
Then, before starting the next epoch, we tune the parameters to \(\boldsymbol{\theta_t'}\). The hyper-parameter \(\lambda\) controls the 
trade-off between previous and current parameters. 

The value of \(\lambda\) is critical to \name.
In the early stage of AT, the model has not yet fit the training data well enough. Thus, the model is not robust enough against adversarial attacks, and its parameters are not very informative. 
If \(\lambda\) is too large, the model mainly relies on previous parameters and learns the training data in small steps, leading to slow convergence.
Therefore, \(\lambda\) should be small in the early stage. 

As the training continues, the model starts to learn enough information from the adversarial examples and attains adversarial robustness. 
In the later stage, the model has learned enough information and gained good robustness. 
\(\lambda\) should be close to 1, otherwise the model tends to discard useful information learned over the training process and overfit the current adversarial examples.



According to the above analysis, \(\lambda\) should change over the course of training, instead of using a fixed value. 
The value of \(\lambda\) should be small in the early stage of training and gradually increase along with the training, which not only ensures the convergence speed but also alleviates the overfitting issue in AT. In this paper, we set \(\lambda\) as follows:
\begin{equation}
\label{eq5}
    \lambda=g(n)=\frac{n+1}{n+c}, \quad  c \geq 1 ,
\end{equation}
where \(n\) denotes the current number of training epochs. \(c\) is a hyper-parameter and we set \(c = 10\) in this work. The accuracy curves of different \(\lambda\) values are as shown in Figure~\ref{fig:MOMAT_test_accuracy}. We will verify that a dynamic \(\lambda\) that varies with the number of training epochs has better robustness against a fixed value in Section~\ref{sec:furtherStudy}.

Algorithm \ref{alg: MOMAT} concludes the overall framework.
Since \name does not restrict the type of loss function in the framework, it is flexible and can be combined with various adversarial training methods such as TRADES~\cite{Trade-off}, MART~\cite{MART} and GAIRAT~\cite{GAIRAT}. 

\begin{table*}[!htp]
\caption{The accuracy (\%) of our method and AT baselines under various adversarial attacks on CIFAR10, CIFAR100 and SVHN datasets with ResNet18 model. 
}
\label{table 1: Combination with other defense methods}
\begin{center}
\begin{small}
\begin{sc}
\begin{tabular}{c|c|cccccc}
\toprule
Dataset & Method & NAT & PGD$^{20}$ & PGD$^{100}$ & MIM & CW & AA \\
\midrule
\multirow{6}{*}{CIFAR10}
&PGD-AT    &\textbf{84.28}& 50.29& 50.12 &51.21&49.31&46.33 \\
&TRADES    & 82.39 & 53.60& 53.65 &54.55&50.90&48.04 \\
&MART     & 81.91& 53.70& 53.70 &54.95&49.35&47.45\\
&GAIRAT       & 81.69& \textbf{55.84}& \textbf{55.90} &\textbf{56.62}&45.50&40.85 \\
\cline{2-8}
&\name & 79.08& 51.81& 51.74 &52.63&49.32&47.08\\
&\name+NMSE    &80.76& 53.54& 53.59 &54.51&\textbf{51.72}&\textbf{48.80}     \\
\midrule
\multirow{6}{*}{CIFAR100}
&PGD-AT    & 58.48& 28.36& 28.33 &29.30&27.06&23.85 \\
&TRADES  & 57.98& 29.90& 29.88 &29.55&26.14&24.72 \\
&MART &55.26&30.10&30.16&30.51&26.00&23.77\\
&GAIRAT       & 50.26& 23.33& 23.35 &23.90&21.55&19.26 \\
\cline{2-8}
&\name & \textbf{58.84}&29.11&29.14&29.97&27.89&24.15\\
&\name+NMSE    & 54.34&\textbf{31.11}&\textbf{30.99}&\textbf{31.42}&\textbf{28.45}&\textbf{25.79}     \\
\midrule
\multirow{6}{*}{SVHN}
&PGD-AT     & \textbf{93.85}& 59.01& 58.92 &59.93&48.66&43.02\\
&TRADES & 90.88& 59.50& 59.43 &60.52&52.76&46.59\\
&MART &88.73&59.45&59.42&59.83&60.19&44.65\\
&GAIRAT &90.50&54.14&54.09&55.88&50.71&44.57\\
\cline{2-8}
&\name&92.15&59.53&59.61&61.32&55.54&50.93\\
&\name+NMSE & 91.70& \textbf{61.21}& \textbf{61.43} &\textbf{61.97}&\textbf{55.88}&\textbf{51.29}\\
\bottomrule
\end{tabular}
\end{sc}
\end{small}
\end{center}
\centering
\end{table*}

\subsection{The NMSE Regularization}
According to the discussion in Section~\ref{section:3}, instead of aligning the clean and adversarial examples by classification probabilities, we utilize the output logits normalized with \(l_2\)-norm.

We align the clean and adversarial examples by minimizing the mean square error between their normalized output logits. Besides, we set \((1-p_{clean})\) as the weight for different adversarial examples so that the model will pay more attention to the clean examples which are vulnerable. We formulate the Normalized Mean Square Error (NMSE) regularization as follows:
\begin{equation}
    \mathcal{L}_{NMSE}=(1-p_{clean})\cdot
    \left\|\frac{f_{\boldsymbol{\theta}}(\mathbf{x})}{||f_{\boldsymbol{\theta}}(\mathbf{x})||_{2}}-\frac{f_{\boldsymbol{\theta}}(\mathbf{x}')}{||f_{\boldsymbol{\theta}}(\mathbf{x}')||_{2}}\right\|_2^2,
\end{equation}
where \(\mathbf{x}'\) is the adversarial example, \(f_{\boldsymbol{\theta}}(\mathbf{x})\) is the output logits of the model, and \(||\cdot||_2\) denotes \(l_2\)-norm of vector.


In summary, the overall loss function in \name framework with NMSE is as follows:
\begin{equation}
    \mathcal{L}= \mathcal{L}_{CE}+ \mu \cdot \mathcal{L}_{NMSE}, 
    \label{Total Loss}
\end{equation}
where \(\mu\) is a hyper-parameter to 
trade off 
the cross-entropy loss \(\mathcal{L}_{CE}\) on adversarial examples and the NMSE regularization term \(\mathcal{L}_{NMSE}\).

\section{Experiments}
\label{sec:result}

In this section, we conduct experiments on several benchmark datasets to evaluate the defense efficacy of our \name framework and NMSE regularization. We also verify that \name could boost the model robustness when combined with various AT methods. 

\subsection{Experimental Setup}
\textbf{Datasets and Models} \quad
We conduct experiments on three benchmark datasets including CIFAR10~\cite{CIFAR100}, CIFAR100~\cite{CIFAR100}, and SVHN~\cite{SVHN}. All images are normalized into \([0, 1]\). We evaluate on two models, ResNet18~\cite{Resnet18} and WRN-32-10~\cite{WRN}, to verify the efficacy of 
our method. 

\textbf{Evaluation Details} \quad
We compare the \name combined with NMSE regularization with the following AT baselines: TRADES~\cite{Trade-off}, MART~\cite{MART}, and GAIRAT~\cite{GAIRAT}. To thoroughly evaluate the defense efficacy
of our method and the baselines, we adopt various adversarial attacks including PGD~\cite{PGD}, MIM~\cite{MIM}, CW~\cite{CW}, and AA~\cite{AA}.

\textbf{Training Details} \quad
For all the experiments, we train ResNet18 (WRN-32-20) using SGD with 0.9 momentum for 120 (180) epochs. The weight decay is $3.5\times10^{-3}$ for ResNet18 and $7\times 10^{-4}$ for WRN-32-10 on three datasets. The initial learning rate for ResNet18 (WRN-32-10) is 0.01 (0.1) till epoch 60 (90) and then linearly decays to 0.001 (0.01), 0.0001 (0.001) at epoch 90 (135) and 120 (180).
To address the cold boot problem of training, we perform standard training on the clean data for the first 10 epochs, and then perform adversarial training. For crafting adversarial examples, the maximum perturbation of each pixel is $\epsilon=\frac{8}{255}$ with the PGD step size $\kappa=\frac{2}{255}$ and step number of 10. 
For the baseline of TRADES, we adopt $\beta$ = 6 for the best robustness.

\begin{table}
\caption{The accuracy (\%) of our method and AT baselines under adversarial attacks on CIFAR10 and CIFAR100 datasets with WRN-32-10 model. 
}
\label{tab:MOMAT_WRN_Part}
\begin{center}
\begin{small}
\begin{sc}
\resizebox{\columnwidth}{!}{
\begin{tabular}{c|c|ccc}
\toprule
Dataset &Method & NAT & PGD$^{20}$ & AA \\
\midrule
\multirow{5}{*}{CIFAR10}
&PGD-AT     & \textbf{86.87}&48.77&47.78\\
&TRADES    & 82.13&55.14&50.38 \\
\cline{2-5}
&\name    & 85.56&52.80&48.35\\
&\name+TRADES    &82.08&\textbf{58.93}&53.73\\
&\name+NMSE &85.04&58.04&\textbf{53.83}\\ 
\midrule
\multirow{5}{*}{CIFAR100}
&PGD-AT    &59.30&28.13&23.99 \\
&TRADES &57.99&31.97&26.76\\ 
\cline{2-5}
&\name    & 60.09&34.46&29.47 \\
&\name+TRADES    & 59.78& 34.52&29.25     \\
&\name+NMSE     & \textbf{61.04}&\textbf{35.15}&\textbf{30.07} \\
\bottomrule
\end{tabular}
}
\end{sc}
\end{small}
\end{center}
\centering
\end{table}

\begin{table}[t]
\caption{The accuracy (\%) of AT methods with or without NMSE regularization under adversarial attacks on CIFAR10 and CIFAR100 dataset with ResNet18 model.
}
\label{table 3: NMSE_table_part}
\centering
\begin{center}
\begin{small}
\begin{sc}
\resizebox{\columnwidth}{!}{
\begin{tabular}{c|c|ccc}
\toprule
Dataset &Method& NAT & PGD$^{20}$ & AA \\
\midrule
\multirow{4}*{CIFAR10}
&PGD-AT   &84.28& 50.29&46.33 \\
&PGD-AT+NMSE    & \textbf{84.77}& 51.56&46.60 \\
&\name & 79.08& 51.81&47.08\\ 
&\name+NMSE    & 80.76& \textbf{53.55}&\textbf{48.70}     \\
\midrule
\multirow{4}*{CIFAF100}
&PGD-AT    & 58.48& 28.36&23.85 \\
&PGD-AT+NMSE    & \textbf{58.88}& 29.55&24.82 \\
&\name & 58.84& 29.11&24.15\\
&\name+NMSE    & 54.34&\textbf{31.11}&\textbf{25.79}     \\
\bottomrule
\end{tabular}
}
\end{sc}
\end{small}
\end{center}
\end{table}


\begin{figure*}
    \centering
    \subfigure[CIFAR10]{
    \includegraphics[width=0.3\textwidth]{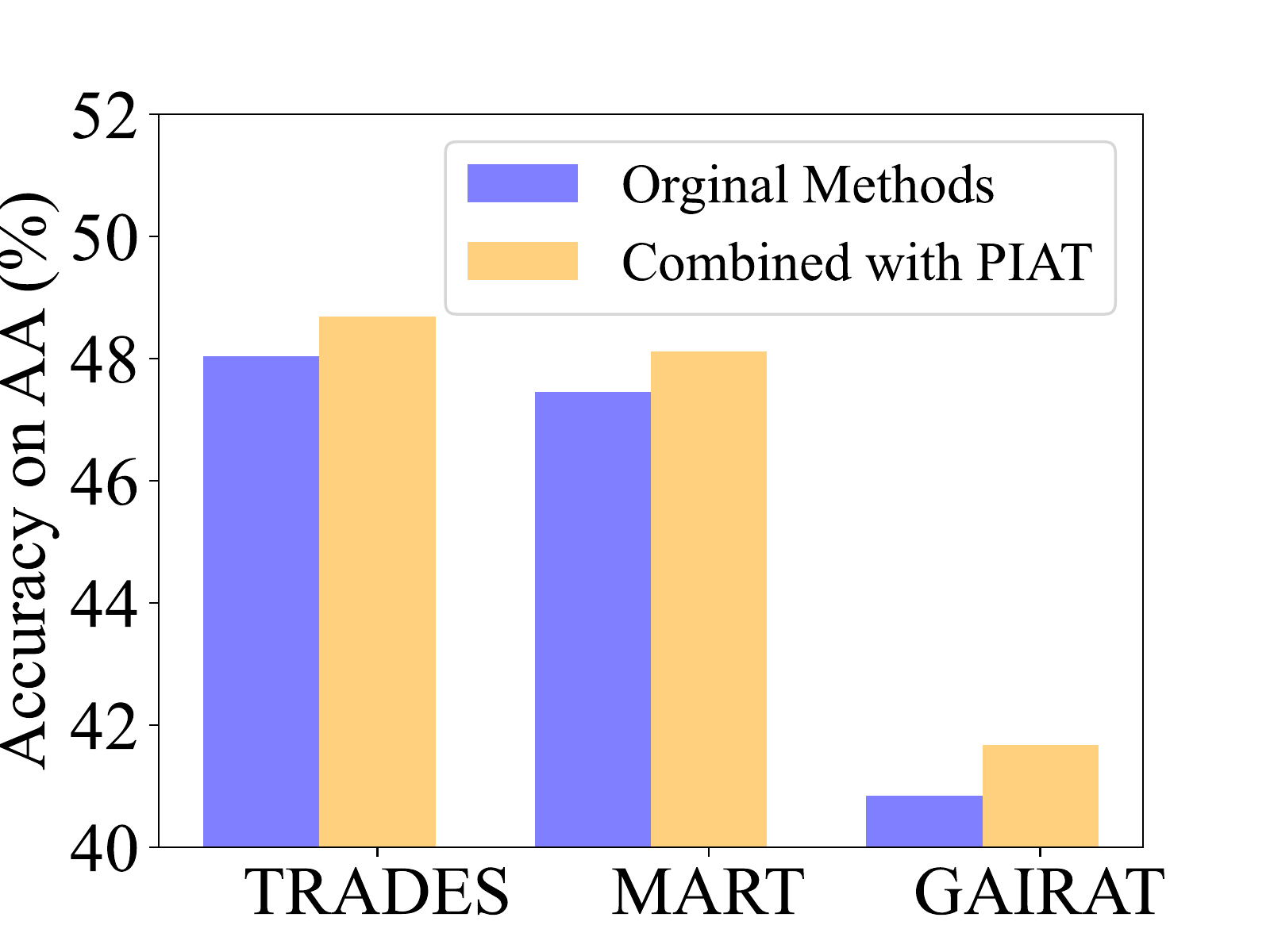}}
    \subfigure[CIFAR100]{
    \includegraphics[width=0.3\textwidth]{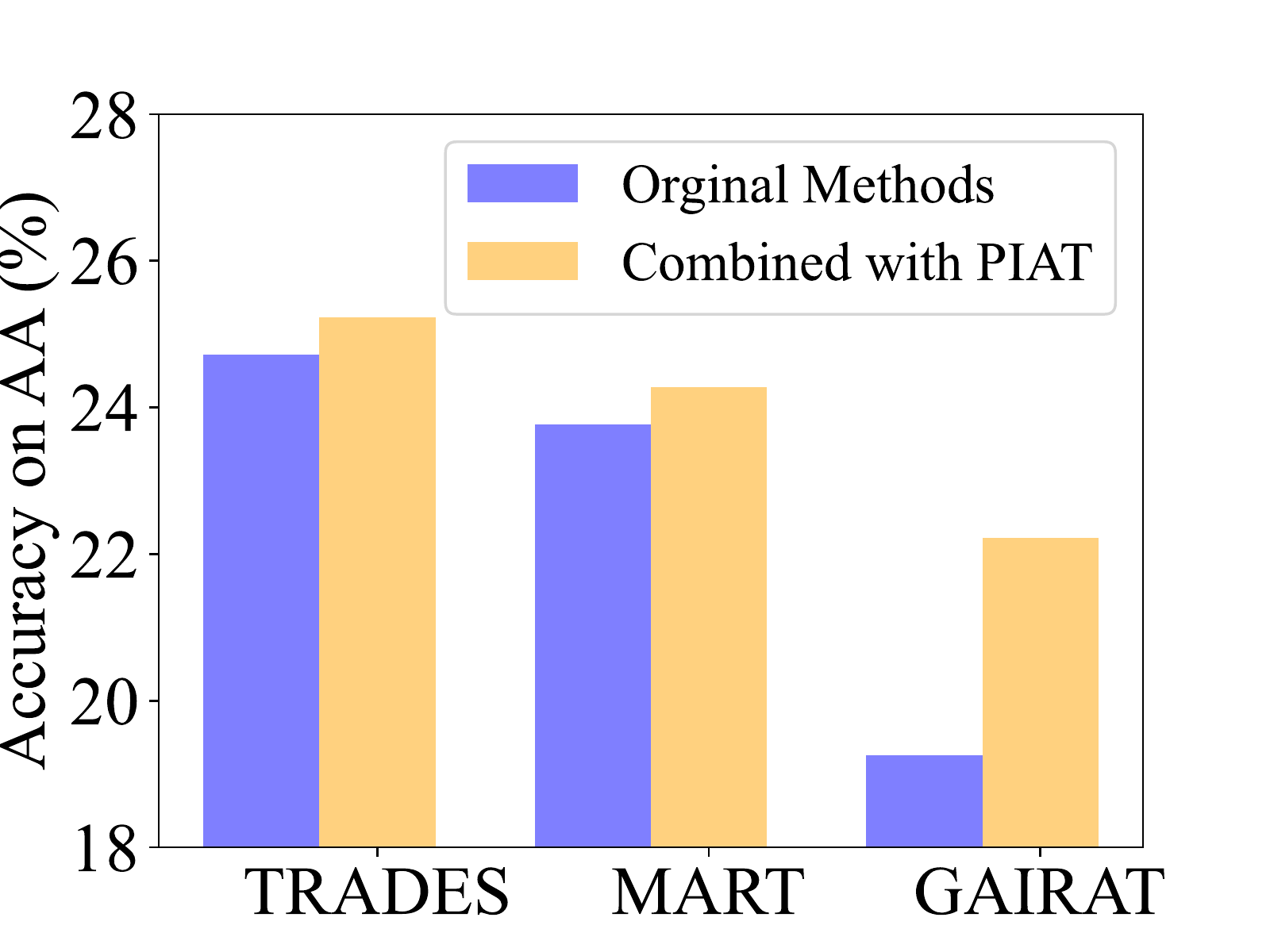}}
        \subfigure[
        SVHN
        ]
        {
    \includegraphics[width=0.3\textwidth]{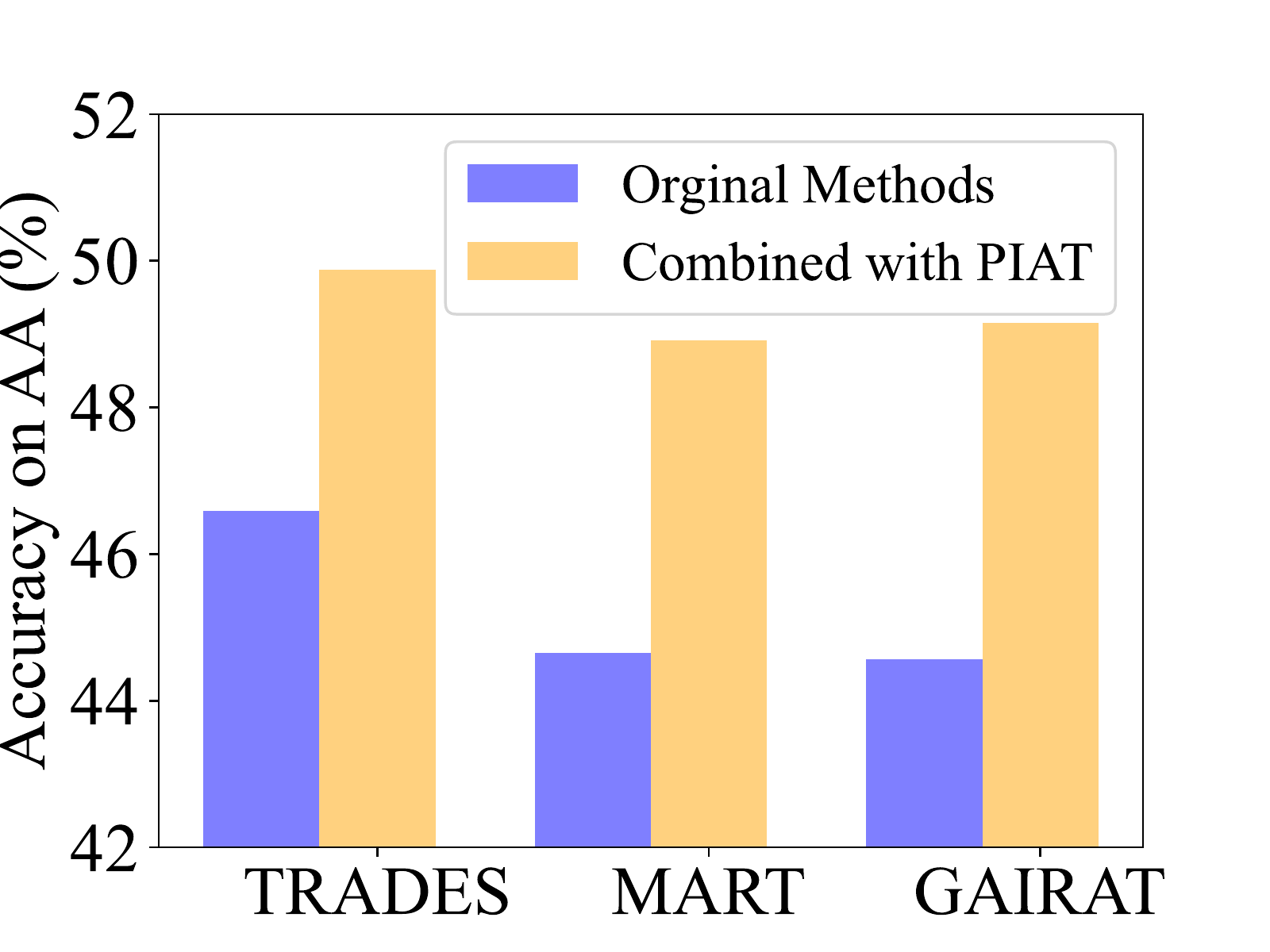}}
    \caption{The accuracy of the \name framework combined with various AT methods under the AA attack on CIFAR10, CIFAR100, and SVHN datasets with ResNet18 model.
    }
    \label{fig:MOMAT combined with classical adversarial training}
\end{figure*}

\subsection{Evaluation on Defense Efficacy}
\label{Result}
We compare the defense efficacy of our method with four AT baselines including PGD-AT, TRADES, MART and GAIRAT. 
Table~\ref{table 1: Combination with other defense methods} reports the accuracy of ResNet18 model trained with our method or the defense baselines under various adversarial attacks on three datasets.

As shown in Table~\ref{table 1: Combination with other defense methods}, our method  (\name + NMSE) exhibits the best performance except for the PGD and MIM attack on CIFAR10 dataset. Under the AA attack, our method achieves 48.80\%, 25.79\% and 51.29\% accuracy on CIFAR10, CIFAR100 and SVHN datasets. Compared to the best results of defense baselines, we gain an improvement of 0.76\%, 1.07\% and 5.50\% on the three datasets, respectively, indicating the great superiority of our method.

We do further evaluation on the WRN-32-10 model, which is much larger than ResNet18, and compare our method with PGD-AT and TRADES. 
The results are reported in Table~\ref{tab:MOMAT_WRN_Part}. Similar to the results on ResNet18, our method achieves the dominant robustness with a clear margin, especially on CIFAR10 dataset.


\begin{figure}
    \vspace{-1em}
    \centering
    {
    \includegraphics[width=\columnwidth]{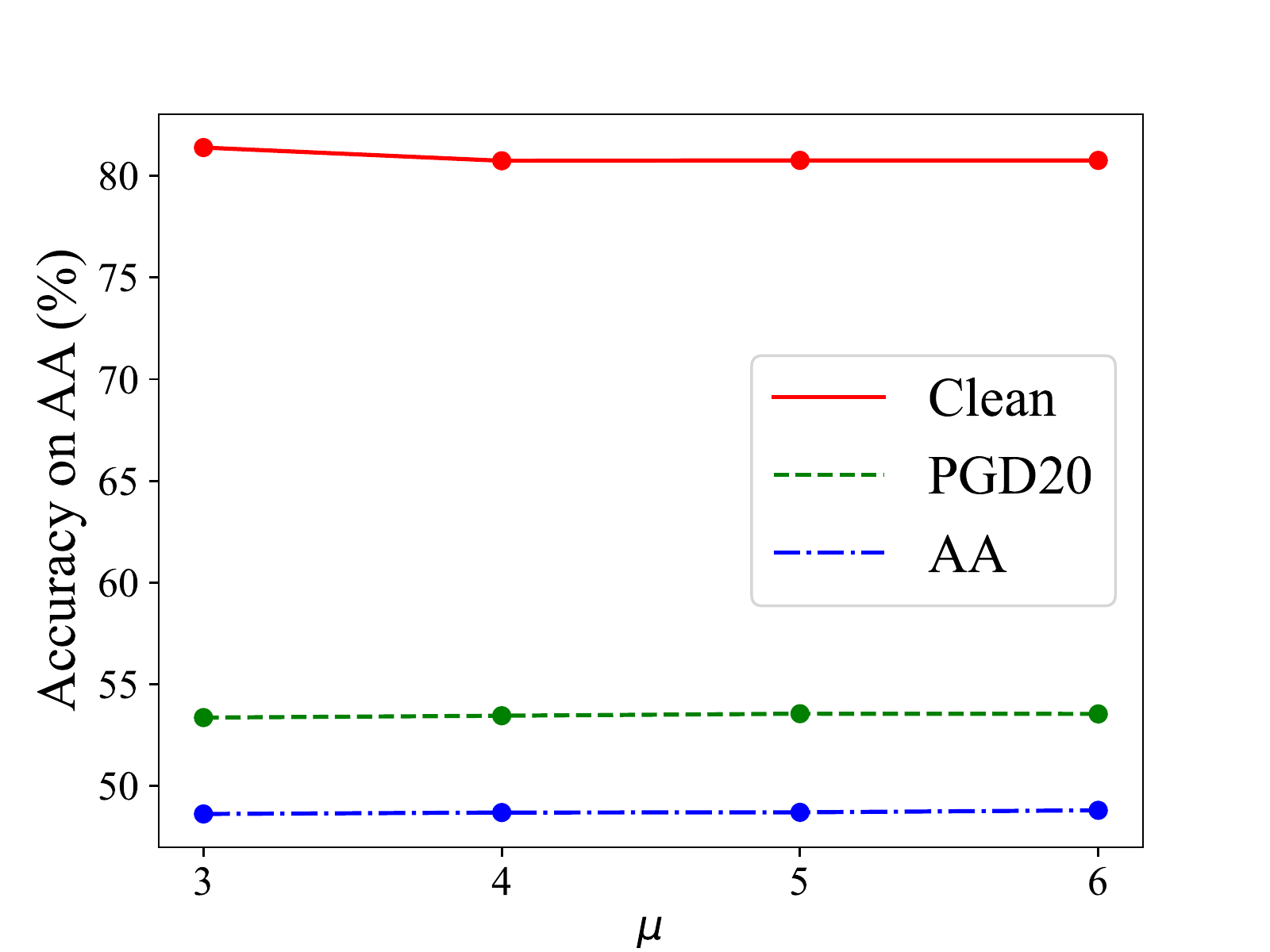}}
    \caption{The clean and robust accuracies of different hyper-parameters of NMSE loss combined with \name framework on CIFAR10. Different 
    lines represent the defensive efficacy of clean and adversarial examples.}
    \label{fig:Hyperparameter of NMSE}
\end{figure}
\begin{figure*}
    \centering
    \subfigure[Loss landscape of PGD-AT]{
    \includegraphics[width =\columnwidth]{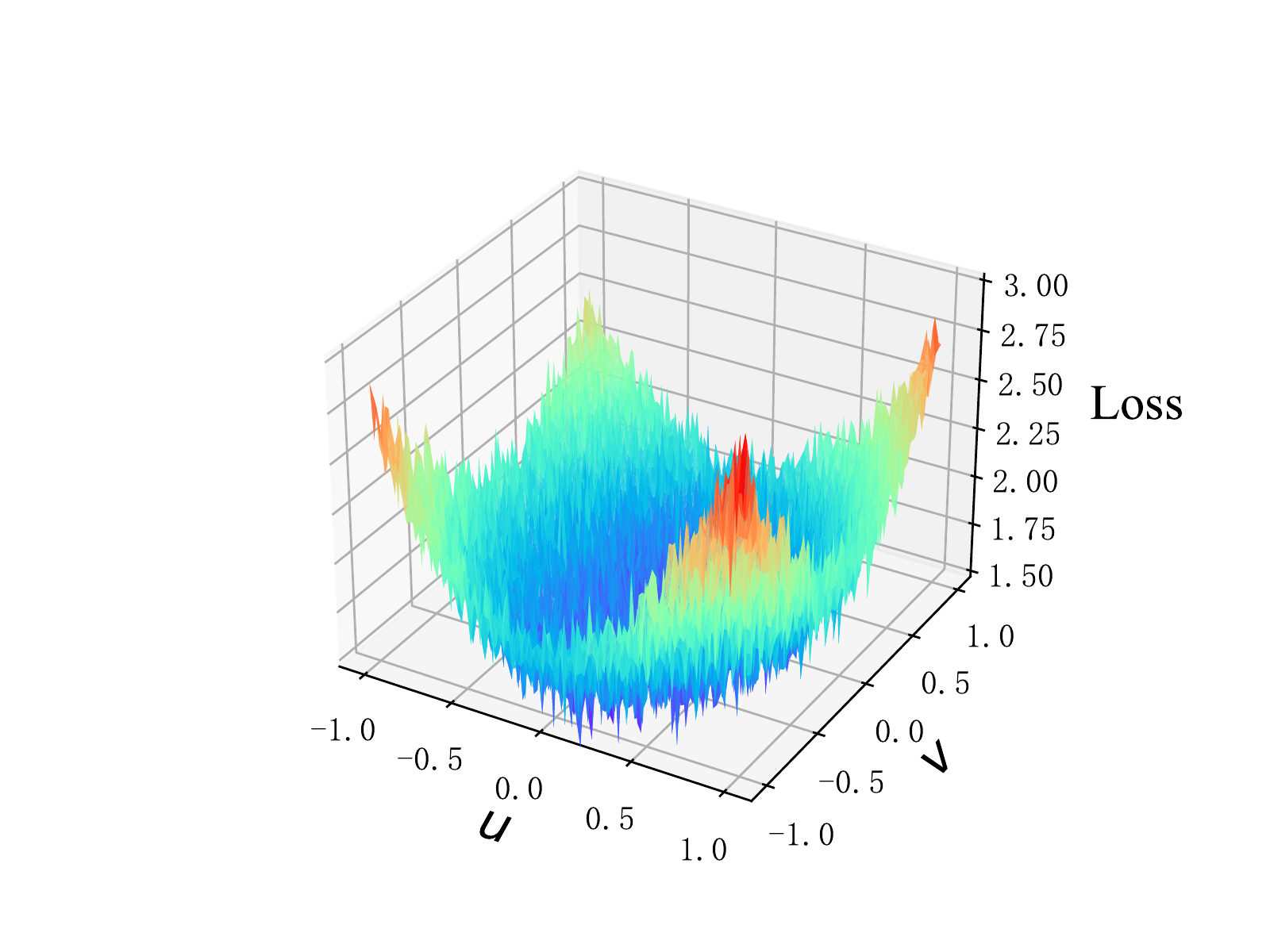}}
    \subfigure[Loss landscape of \name]{
        \includegraphics[width=\columnwidth]{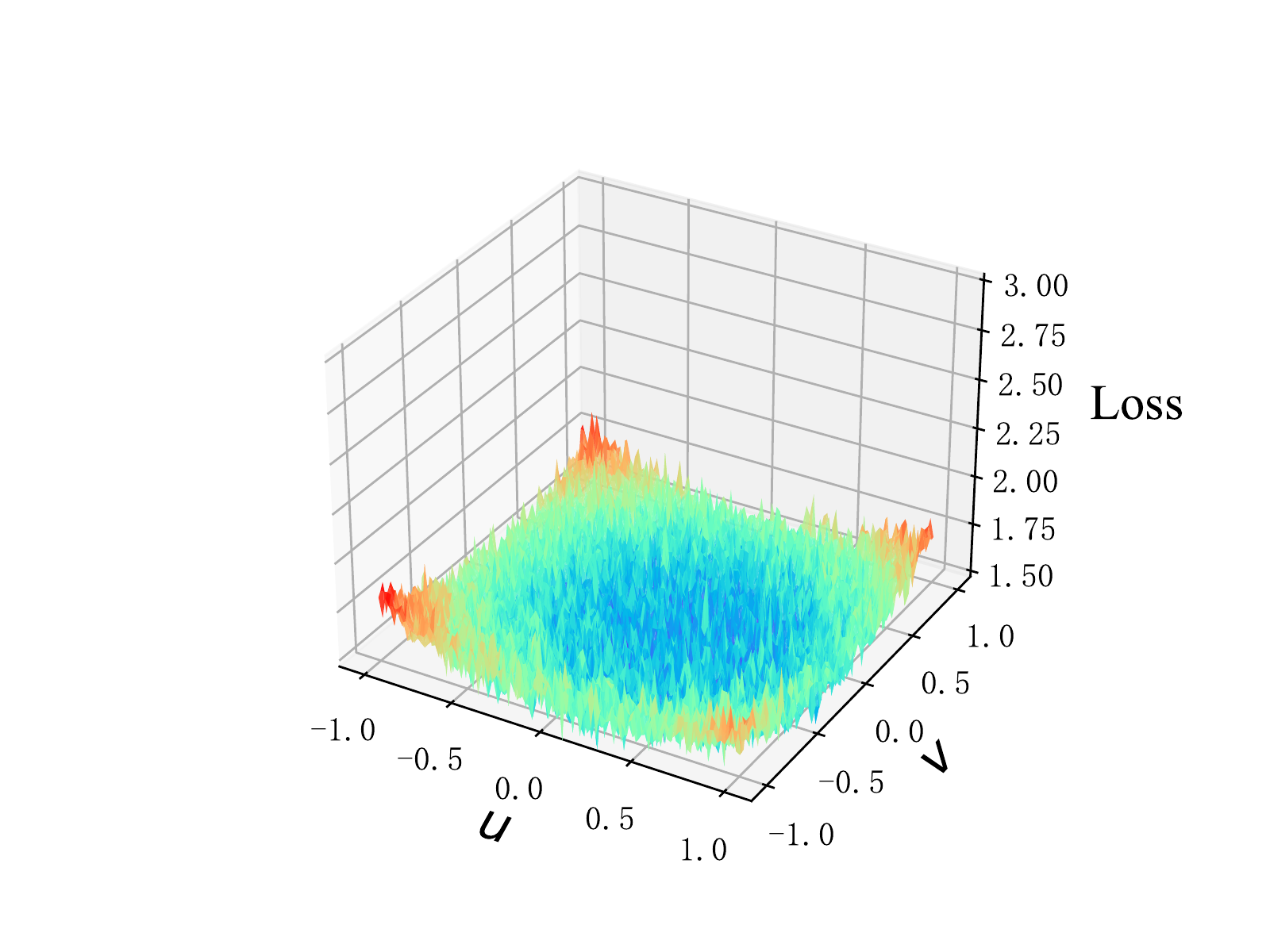}}
    \caption{Illustration of the loss landscape in three dimensions. Our \name framework has flatter loss landscape than PGD-AT.
    }
    \label{fig:MOMAT model loss}
\end{figure*}

\subsection{Ablation Study}

\textbf{\name Framework} \quad
Since \name is a general framework, we 
incorporate other AT methods into \name 
to demonstrates its defense efficacy. 
Figure~\ref{fig:MOMAT combined with classical adversarial training} illustrates the accuracy of \name framework combined with TRADES, MART, and GAIRAT, respectively, under the AA attack on the three datasets. 

As shown in Figure~\ref{fig:MOMAT combined with classical adversarial training}, over all the three datasets, \name boosts the robustness of various AT methods against the AA attack. The results demonstrate that we can easily 
 incorporate other AT methods into our \name framework 
without incurring any additional cost to achieve better robustness and performance. 


We also 
do ablation study 
on the WRN-32-10 model and report the results in Table~\ref{tab:MOMAT_WRN_Part}. 
Specifically, \name framework significantly boosts the robust accuracy of TRADES, gaining absolute improvement of $3.35\%$ and $2.49\%$ on CIFAR10 and CIFAR100 under the AA attack, respectively. 
Our framework leads to higher robust accuracy when combined with other AT methods on two models, indicating that \name has good flexibility and generalization.

\textbf{NMSE Regularization} \quad
To evaluate the effectiveness of our proposed NMSE regularization, we observe the performance of PGD-AT and \name with or without NMSE regularization. Table~\ref{table 3: NMSE_table_part} reports the accuracy of ResNet18 models against PGD and AA attack on the CIFAR10 and CIFAR100 datasets.

As shown in Table~\ref{table 3: NMSE_table_part},  
both PGD-AT and \name achieve better robustness with NMSE regularization than that without NMSE. For instance, the accuracy of PGD-AT against AA attack gains absolutely 0.37\% and 1.62\% on CIFAR10 and CIFAR100, respectively. It indicates that the NMSE regularization greatly helps the model learn the features of adversarial examples. 



\subsection{Further Analysis}
\label{sec:furtherStudy}
We continue to analyze the sensitivity of hyper-parameters in our method. 
Besides, we draw a 3D visual landscape of loss function to further compare our method and PGD-AT. 


\textbf{Hyper-parameter Analysis} \quad
The hyper-parameter \(\lambda\) in Eq.~\ref{eq:MOMAT} is used to control the trade-off between previous and current parameters for \name. In Section~\ref{section:3}, we intuitively suggest that \(\lambda\) should change over the course of training, instead of using a fixed value. 
To verify this point, we compare the accuracy of \name combined with NMSE using fixed $\lambda=0,0.1,0.3,0.5,0.7,0.9$ and our variable \(\lambda\) as in Eq.~\ref{eq5}. Figure~\ref{fig:MOMAT_test_accuracy} illustrates the results on the CIFAR10 dataset. It is obvious that the variable strategy of  \(\lambda\) is vital to alleviate the oscillations in the early stage and the overfitting issue in the later stage of the AT process. 

The hyper-parameter \(\mu\) in Eq.~\ref{Total Loss} is used to trade off the cross-entropy loss on adversarial examples and the NMSE regularization term. We study the accuracy of \name combined with NMSE using different \(\mu\). 
Figure~\ref{fig:Hyperparameter of NMSE} illustrates the results on the CIFAR10 dataset when we take $\mu=3,4,5,6$.
It indicates that the defense efficacy of our method is not sensitive to $\mu$. 
Similar observations can be obtained on the CIFAR-100 dataset.
Therefore, we set $\mu=5$ in our experiments for 
an appropriate 
trade-off between the accuracy on clean and adversarial examples. 

\textbf{Loss Landscape} \quad
To comprehensively evaluate the efficacy of our framework,  
we refer to the method proposed by Li \etal~\cite{visual_network} and compare the model obtained by \name framework and PGD-AT in 3D. 
Let $\mathbf{u}$ and $\mathbf{v}$ be two random direction vectors sampled from the Gaussian distribution. We plot the loss landscape around $\boldsymbol{\theta}$ of the following equation when inputting the same data:
\begin{equation}
    \mathcal{L}(\boldsymbol{\theta};\mathbf{u};\mathbf{v})=\mathcal{L}\left(\boldsymbol{\theta}+m_1\frac{\mathbf{u}}{||\mathbf{u}||}+m_2\frac{\mathbf{v}}{||\mathbf{v|}|}\right),
\end{equation}
where \(m_1,m_2 \in [-1,1]\).

Figure~\ref{fig:MOMAT model loss} illustrates the shape of the 3D landscape map. We observe that compared with the PGD-AT, the model trained using the \name framework exhibits less fluctuation in the loss landscape under the same perturbation.
Compared with the PGD-AT, the 3D landscape obtained using the \name framework indicates that the model converges to a flatter area and has better robust accuracy.

\section{Conclusion}
In this work, we propose the \name framework to make the decision boundary change more moderately, thus eliminating the oscillation phenomenon in the early stage of AT. Also, in the later stage, \name framework considers the decision boundary of previous iteration, preventing the decision boundary from becoming too complex and alleviating the overfitting issue. Moreover, we suggest to use the Normalized Mean Square Error (NMSE) as regularzation to align the clean examples and adversarial examples, that focuses more on the relative magnitude of the output logits rather than the absolute magnitude.
Extensive experiments verify that our framework can eliminate the oscillation phenomenon and alleviate the overfitting issue of adversarial training. Furthermore, the \name combined with NMSE loss could significantly improve the model robustness of adversarial training without extra computational cost.
In addition, our framework is flexible and general, and various adversarial training methods can be combined into our \name framework to further boost their performance.  

Our work shows that the historical information of adversarial training process is very useful. We hope our work will inspire more works utilizing historical training information to boost the model robustness.



{\small
\bibliographystyle{ieee_fullname}
\bibliography{egbib}
}

\end{document}